\documentclass[runningheads]{llncs}

\usepackage{eccv}

\usepackage{eccvabbrv}
\usepackage{adjustbox}
\usepackage{pifont}
\usepackage{amsmath}
\usepackage{graphicx}
\usepackage{booktabs}

\newcommand{\cmark}{\ding{51}}%
\newcommand{\xmark}{\ding{55}}%

\usepackage[accsupp]{axessibility}  

\usepackage[pagebackref,breaklinks,colorlinks,citecolor=eccvblue]{hyperref}
\usepackage{graphicx}

\usepackage{orcidlink}

\newcommand{\dname}{WeatherProof dataset}

\usepackage[pagebackref,breaklinks,colorlinks,citecolor=eccvblue]{hyperref}
\usepackage{graphicx}

\usepackage{orcidlink}

\usepackage{graphicx}
\usepackage{amsmath}
\usepackage{amssymb}
\usepackage{booktabs}
\usepackage{makecell}
\usepackage{multirow} 
\usepackage[ruled,longend]{algorithm2e} 
\usepackage{blindtext}
\usepackage{enumerate}
\usepackage{subcaption}
\usepackage{adjustbox}
\usepackage{pifont}
\usepackage[pagebackref,breaklinks,colorlinks]{hyperref}

\usepackage[pagebackref,breaklinks,colorlinks]{hyperref}

\usepackage[capitalize]{cleveref}
\crefname{section}{Sec.}{Secs.}
\Crefname{section}{Section}{Sections}
\Crefname{table}{Table}{Tables}
\crefname{table}{Tab.}{Tabs.}

\begin{document}

\title{WeatherProof: Leveraging Language Guidance for Semantic Segmentation in Adverse Weather} 

\titlerunning{Abbreviated paper title}

\makeatletter
\newcommand{\printfnsymbol}[1]{%
  \textsuperscript{\@fnsymbol{#1}}%
}
\makeatother

\author{Blake Gella\thanks{Equal contribution.} \inst{1} \and
Howard Zhang\printfnsymbol{1} \inst{1} \and
Rishi Upadhyay\inst{1} \and
Tiffany Chang\inst{1} \and
Nathan Wei\inst{1} \and
Matthew Waliman\inst{1} \and
Yunhao Bao\inst{1} \and
Celso de Melo\inst{3} \and
Alex Wong\inst{2} \and
Achuta Kadambi\inst{1}}

\authorrunning{F.~Author et al.}

\institute{UCLA \and
Yale University \and
DEVCOM Army Research Laboratory}

\maketitle

\begin{abstract}
  We propose a method to infer semantic segmentation maps from images captured under adverse weather conditions. We begin by examining existing models on images degraded by weather conditions such as rain, fog, or snow, and found that they exhibit a large performance drop as compared to those captured under clear weather. To control for changes in scene structures, we propose WeatherProof, the first semantic segmentation dataset with accurate clear and adverse weather image pairs that share an underlying scene. Through this dataset, we analyze the error modes in existing models and found that they were sensitive to the highly complex combination of different weather effects induced on the image during capture. To improve robustness, we propose a way to use language as guidance by identifying contributions of adverse weather conditions and injecting that as ``side information''. Models trained using our language guidance exhibit performance gains by up to 10.2\% in mIoU on WeatherProof, up to 8.44\% in mIoU on the widely used ACDC dataset compared to standard training techniques, and up to 6.21\% in mIoU on the ACDC dataset as compared to previous SOTA methods.
\end{abstract}

\section{Introduction}
\label{sec:intro}
Semantic segmentation has a rich history due to its countless applications in autonomous driving~\cite{ess2009segmentation,nekrasov2019real,siam2018rtseg,zhao2018icnet}, robotics~\cite{kim2018indoor,milioto2018real,milioto2019bonnet}, and scene understanding~\cite{gupta2015indoor,cordts2016cityscapes,li2009towards}. Current state-of-the-art model architectures have achieved high scores, i.e. mean intersection-over-union (mIoU), on competitive semantic segmentation benchmarks like ADE20K~\cite{zhou2017scene} or Cityscapes~\cite{cordts2016cityscapes}. Yet, despite their success on these leaderboards, when presented with images with visual degradations, i.e., those captured under adverse conditions, their performance similarly degrades.

\begin{figure}[tb]
    \centering
    \includegraphics [width=\linewidth]{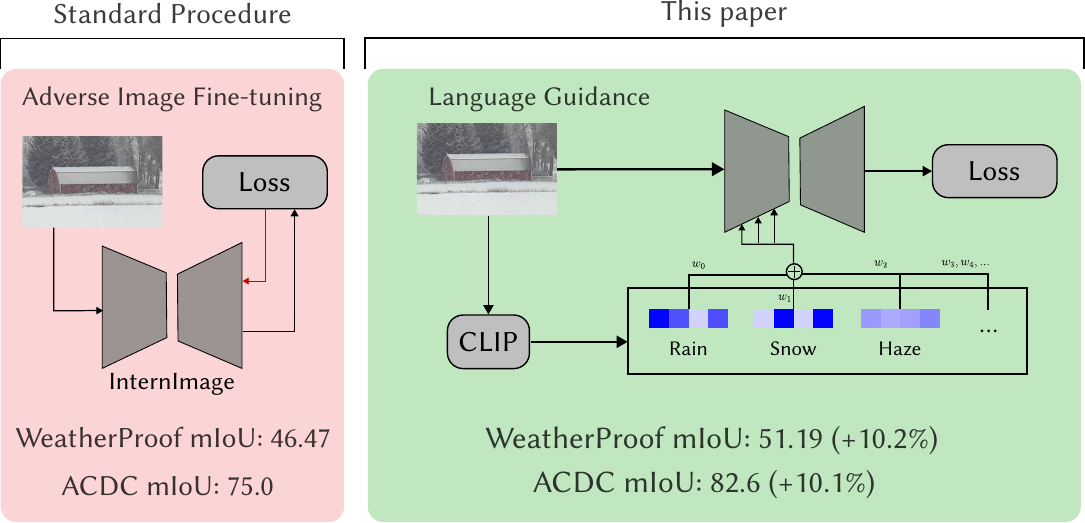}
    \caption{\textbf{By leveraging CLIP-based language guidance, our models perform up to 10.2\% better on our WeatherProof test set, and 8.4\% better on the widely used ACDC dataset as compared to standard fine-tuning procedures.}}
    \label{fig:teaser}
\end{figure}


The real-world effects of adverse weather conditions such as rain, fog, or snow have been shown by many previous studies to have very complex visual degradation patterns\cite{wang2019spatial,ba2022not,zhang2023weatherstream,liu2018desnownet,tan2008visibility} -- patterns that can be affected by factors including but not limited to atmospheric condition, camera parameters, or even geographic location\cite{zhang2023weatherstream,ba2022not}. The prevalence of weather in our natural world translates to performance gaps that arise in our algorithms. Existing work~\cite{sakaridis2018semantic,sakaridis2021acdc,kerim2022semantic, tung2017raincouver, shaik2024idd, yu2020bdd100k} has provided datasets and methods with the goal of studying the effects of these natural phenomena. However, as it is difficult to capture paired datasets to study them in a controlled setting, existing datasets have resorted to the use of synthetic rendering of weather onto the image to test their algorithms, or are limited to data that include mis-alignments in the underlying scene between adverse and clear-weather images. To address this, we build off the WeatherStream dataset~\cite{zhang2023weatherstream} to introduce the \dname, the first semantic segmentation dataset with accurately paired clear and weather-degraded image pairs. By ensuring the underlying semantic labels are the same between clear and adverse weather images, we provide a controlled test bench where performance degradations can be largely isolated to weather artifacts.

Using this dataset, we tested several recent semantic segmentation model architectures, which underperformed on adverse-weather images compared to clear-weather images (see~\cref{table:Degraded-table}). One common reason for this underperformance is the complex combination of weather effects present in the real world, as seen in \cref{fig:weather_combos}. For instance, scenes with rain frequently include both rain and fog due to a number of physical parameters, like temperature and humidity, which affect the appearance of objects populating the image. The presence of composite blends of weather conditions can lead to highly varying perturbations of the image, which in turn lead to the aforementioned performance gap.

To address this issue, we focus the model on learning weather degradations by injecting ``side information'' regarding compositions of weather effects into the model. Knowledge of weather composition could be beneficial to the model for becoming resilient to weather by serving as an inductive bias to narrow down the solution space. To do this, we propose a CLIP Injection Layer, which does a weighted sum of text encodings of various weather effects and their severity levels and uses cross attention to inject this information into the model.

By biasing the predictions through the use of weather condition compositions, our method obtains relative increases of up to 10.2\% mIOU as compared to standard training procedures on our new \dname\ evaluation set. To further validate the benefits of our method, we evaluate on the commonly used ACDC dataset~\cite{sakaridis2021acdc}, and found that our method improves upon standard training techniques by 8.44\% and previous SOTA methods by 6.21\%. We further show that including the language guidance not only improves the robustness of models against natural weather conditions, but man-made conditions like smoke as well. We show this by evaluating on the A2I2-Haze dataset~\cite{narayanan2023multi}, in which we outperformed standard training procedures by 3.9\%. Code and data will be released to ensure reproducibility.

\subsection{Contributions}

In summary, we make the following contributions:
\begin{itemize}
    
\item We augment the standard model training process through a CLIP Injection Layer, which helps by injecting the composition of the weather effect through cross attention and language guidance, improving performance on adverse weather conditions. 

\item We introduce the \dname -- a semantic segmentation dataset with over 174.0K images. It is the first with high quality semantic segmentation labels, with accurately aligned and paired clear and adverse weather images for more accurate evaluation under controlled settings.

\item We validate our method by improving on adverse weather-degraded images of up to 10.2\% on the \dname, 3.9\% on the A2I2-Haze Dataset~\cite{narayanan2023multi}, and 8.44\% compared to standard training techniques on the ACDC Dataset, and 6.21\% compared to SOTA on the ACDC Dataset \cite{sakaridis2021acdc}.
    
\end{itemize}

\section{Related Works}
\label{sec:related}
\subsection{Semantic Segmentation}
Recently, research has shown the superb performance of deep learning models such as CNNs~\cite{hong2015decoupled,hu2018learning,lin2017refinenet,long2015fully, chen2017deeplab, chen2018encoder, badrinarayanan2016segnet} or vision transformers~\cite{xie2021segformer,hu2021istr}. SegNet uses the pooling indices calculated in the encoder to perform nonlinear upsampling in the decoder, improving the memory and inference efficiency~\cite{badrinarayanan2016segnet}. DeepLab made many advancements regarding extracting information at different scales including using dilated convolutions, Atrous Spacial Pyramid Pooling, and image level features~\cite{chen2017deeplab, chen2018encoder}. With the rising popularity of these convolution and attention-based architectures, a recent wave of research has shown the excellent learning capability and segmentation performance of large, generally pre-trained models. The Swin architecture uses a shifted window technique to achieve both local and global attention while maintaining linear computational complexity with respect to image size~\cite{liu2021swin, liu2022swin}. The ConvNeXt model makes multiple changes to the standard convolutional network training pipeline (kernel sizes, activations, etc.) to modernize the CNN approach to outperform  Swin~\cite{liu2022convnet}. The InternImage model uses deformable convolutions to maintain the long-range dependence of attention layers in a low memory/computation regime~\cite{wang2023internimage}. However, all of these aforementioned models benchmark their segmentation performance on the ADE20K dataset~\cite{zhou2017scene}, which does not contain images in adverse conditions, such as adverse weather. As such, the robustness of the current top performers in the semantic segmentation task to adverse weather are yet to be evaluated.

\begin{table}[tb]
\caption{\textbf{\dname\ is the first high quality annotated segmentation dataset with accurate clear and weather-degraded image pairs for better consistency loss in training and evaluation.} Other datasets either do not contain adverse weather effects, have synthetic weather effects, or do not have accurate paired clear images. While ACDC does have paired images, there exists a stark difference between the adverse and reference images, which is shown in~\cref{fig:dataset}. Similarly, the IDD-AW dataset has paired images but do not offer normal RGB Pairing, but rather, RGB to Near-Infrared Pairing.}
    \scriptsize
    \centering
    \begin{tabular}{ccccc}
    \toprule
    Dataset & $\#$ Images & Weather Degradations? & Real? & Paired? \\
    \midrule
    ADE20K~\cite{zhou2017scene} & 27.5K & \xmark & \cmark & \xmark \\
    Cityscapes~\cite{cordts2016cityscapes} & 25K & \xmark & \cmark & \xmark \\
    Foggy Cityscapes~\cite{sakaridis2018semantic} & 35K & \cmark & \xmark & \cmark \\
    ACDC~\cite{sakaridis2021acdc} & 4K & \cmark & \cmark & \xmark \\
    IDD-AW ~\cite{shaik2024idd} & 5K & \cmark & \cmark & \xmark \\
    BDD100K ~\cite{yu2020bdd100k} & 10K & \cmark & \cmark & \xmark \\
    Raincouver ~\cite{tung2017raincouver} & 326 & \cmark & \cmark & \xmark \\
    \midrule
    \dname\ (Ours) & 174.0K & \cmark & \cmark & \cmark \\
    \bottomrule
    \end{tabular}
\label{tab:related_work}
\end{table}


\subsection{Semantic Segmentation in Adverse Weather}
\label{sec:related_adverse}
The two most popular semantic segmentation datasets are ADE20K~\cite{zhou2017scene} and Cityscapes~\cite{cordts2016cityscapes}. ADE20K does not include images with adverse conditions, and Cityscapes avoids adverse weather conditions as well. Several datasets have since been made to include weather effects~\cite{tung2017raincouver, shaik2024idd, sakaridis2018semantic, yu2020bdd100k, sakaridis2021acdc}. The Raincouver dataset includes 326 hand-annotated images which only consist of rainy images without a clear, non-degraded counterpart for comparison~\cite{tung2017raincouver}. The BDD100K dataset offers a variety of different weather conditions, but similarly does not include paired, clear images~\cite{yu2020bdd100k}. Efforts have since been made to provide adverse conditions through synthetic means, such as the generation of the Foggy Cityscapes dataset~\cite{sakaridis2018semantic}. However, research in the past has shown that there exists a performance gap when training on synthetic weather conditions~\cite{ba2022not,zhang2023weatherstream}. The IDD-AW dataset does provide a paired, clear counterpart to adverse weather frames, but in the form of NIR (Near-Infrared) images rather than RGB~\cite{shaik2024idd}. The ACDC dataset provides real images with segmentation labels with weather conditions present~\cite{sakaridis2021acdc}. They also provide a paired frame for each adverse frame representing the clear weather version. However, it can be seen in~\cref{fig:dataset} that there are inconsistencies in the pairs that rule out its effectiveness in paired training approaches. A comparison of semantic segmentation datasets in adverse weather can be seen in~\cref{tab:related_work}.

\subsection{Language and Vision}
\label{sec:related_language}
The domains of language and vision have often been separate in machine learning. Recent works have begun to incorporate language into vision models~\cite{radford2021learning,li2022blip,rombach2022high,nichol2021glide}. The BLIP model trains a captioner and filter and uses it on Internet images to achieve high performance results on vision-language tasks~\cite{radford2021learning}. The CLIP model learns a shared latent space between image and text by contrastive pretraining on image-text pairs. CLIP has shown to be fundamental for enabling vision models with language priors, as seen with GLIDE~\cite{nichol2021glide} and Stable Diffusion~\cite{rombach2022high}. Stable Diffusion uses CLIP's text encoder to inject prompts into their UNET's cross attention layers, guiding image generation through text. However, vision models have yet to utilize language for guidance through adverse conditions such as weather.

\section{Dataset and Analysis}
In order to accurately evaluate the performance of semantic segmentation models in the presence of adverse weather, we introduce the \dname. Further details and statistics of this dataset will be given in~\cref{sec:methods_dataset}. We analyze the composition of different and multiple weather conditions on a scene and see its effects on the model performance in~\cref{sec:methods_analysis}.

\begin{figure}[t]
    \centering
    \includegraphics [width=\linewidth]{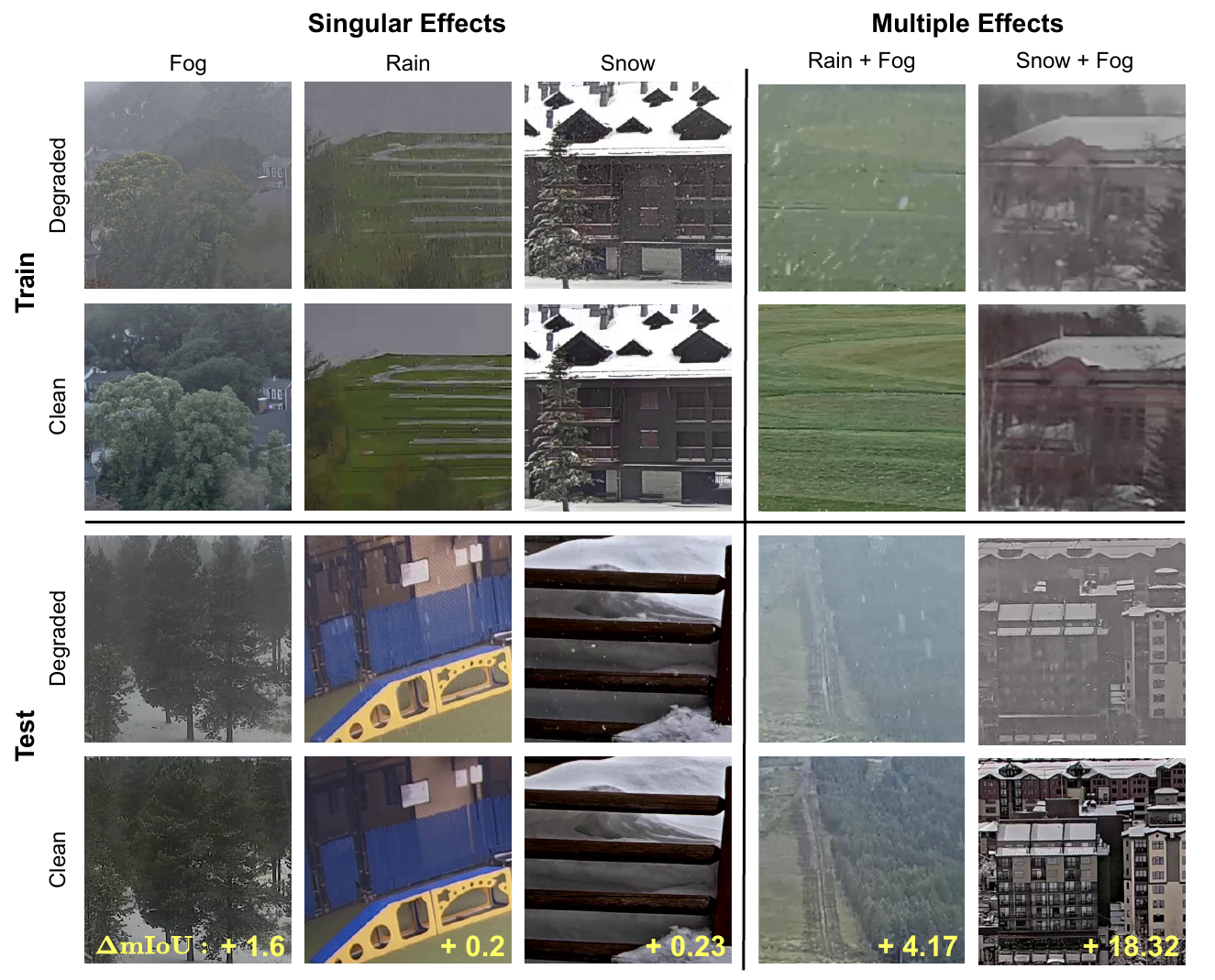}
    \caption{\textbf{The train and test sets of WeatherProof include paired sets of varied combinations of weather effects.} \textit{Top:} Various types of weather effects and their compositions from the training set. \textit{Bottom:} Weather effects and combinations in our test set. Change in mIoU between clear and degraded images of the InternImage baseline is shown in yellow. \textbf{Note the significant impact on mIoU results of multiple combined weather effects.}} 
    \label{fig:weather_combos}
    \vspace{-10pt}
\end{figure}

\subsection{Dataset and Paired Data Testing}
\label{sec:methods_dataset}
\begin{figure}[tb]
    \centering
    \includegraphics [width=\linewidth]{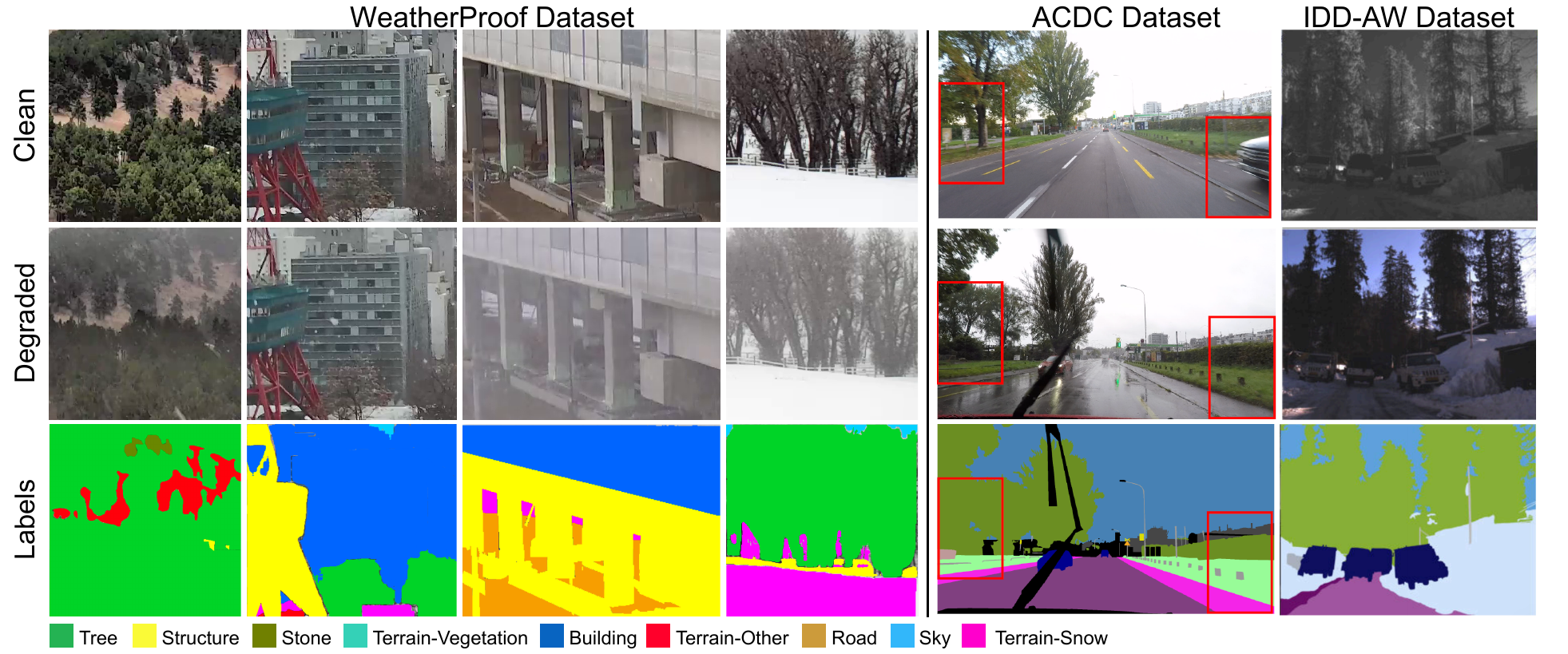}
    \caption{\textbf{\dname\ contains accurate clear and adverse weather image pairs with 10 semantic classes}. The dataset includes rain, snow, and fog weather effects. The labels below the image are for the \dname. In contrast, the ACDC~\cite{sakaridis2021acdc} and IDD-AW~\cite{shaik2024idd} datasets' paired images either have major differences in semantic information and scene structure or are not in RGB space.}
    \label{fig:dataset}
\end{figure}
We train and evaluate our models on our own \dname\ containing 147.8K paired adverse and clear images for training, and 26.2K for testing. The clear and adverse image pairs were selected from the GT-RAIN~\cite{ba2022not} and WeatherStream~\cite{zhang2023weatherstream} datasets. This dataset was chosen due to its meticulous consideration of scene consistency between the clear and degraded images. By leveraging this scene consistency, we are able to construct the first semantic segmentation dataset with truly paired data. Paired data refers to pairs of images captured under clear and adverse condition with minimal differences in the underlying scene (i.e. structural alignment). For example, as~\cref{fig:dataset} depicts, while ACDC and IDD-AW both claim to have ``paired'' data, neither qualify under this definition. ACDC has differences in the underlying scene between clear and degraded frames, and IDD-AW uses NIR rather than RGB data for a paired, clear frame. In our dataset, paired images have the exact same segmentation labels, allowing for better evaluation and analysis of the performance of semantic segmentation models in adverse weather. On the testing side, the paired nature of the \dname\ allows one to get a more accurate evaluation of the exact performance gap that is present due to the specific degradations (i.e. fog, rain and fog, snow, etc.). Since the only differences (apart from extremely minute discrepancies) between two paired images are the presence of weather artifacts, any performance gap between the two is attributed to the adverse weather.

In addition to paired-image evaluation, the WeatherStream dataset also contains geographic locations around the world in a variety of different urban and natural locations, with varying camera parameters and resolutions for dataset variety. We label the 10 following classes in the datasets: background, tree, structure, road, terrain-snow, terrain-grass, terrain-other, stone, building, and sky. To emphasize accurate object borders, we give priority to minimizing the amount of background labels between objects without mislabelling. Samples from the dataset can be seen in~\cref{fig:dataset}.

\subsection{Weather Composition Analysis}
\label{sec:methods_analysis}
\cref{fig:weather_combos} shows examples of weather combinations present in our dataset. In many clear-degraded pairs, we can see the presence of multiple weather effects, notably rain+fog and snow+fog. This is because the same natural conditions (like humidity or temperature) that lead to rain or snow can also lead to the presence of fog. Additionally, previous research has shown that smaller droplets of rain or particles of snow, depending on camera parameters, can lead to a low-frequency haze effect in the scene, which is similar in visual appearance to fog~\cite{zhang2023weatherstream,ba2022not}. 

In~\cref{fig:weather_combos}, we run the test set image pairs through a baseline model (InternImage) that was fine-tuned on the \dname. The mIoU results show that, typically, with more complex weather patterns, i.e., rain+fog or snow+fog, there is a larger performance gap between clear and degraded. This validates that complex weather patterns are more difficult for state-of-the-art models to perform well on as compared to simpler single weather patterns. We introduce the CLIP Injection Layer in the following section to alleviate this issue.

\section{Methods}
\label{sec:clip_methods}
\begin{figure}[tb]
    \centering
    \includegraphics [width=\linewidth]{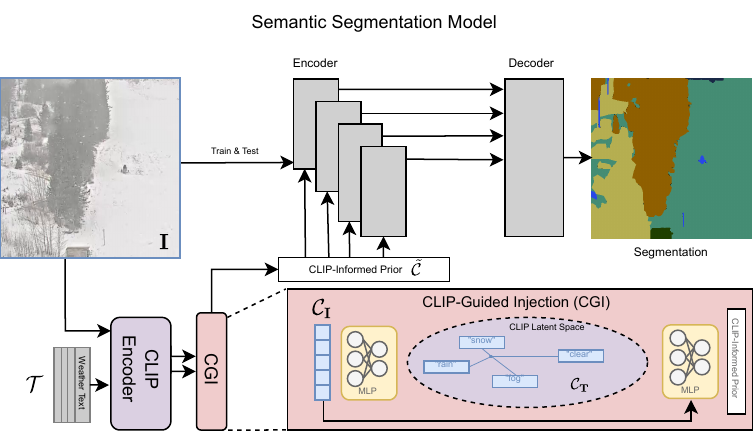}
    \caption{\textbf{By using CLIP-based language guidance, models are able to generate features that are more resilient to adverse weather conditions.} During training, a CLIP-Guided Injection module learns a CLIP-informed prior representing the adverse weather effect in the CLIP latent space. This is concatenated with the image latent before being fed in through cross-attention layers into the model.}
    \label{fig:model}
\end{figure}

 In this section, we introduce and formalize the method used to inject CLIP-based language guidance into our network. See~\cref{fig:model} for an overview of our overall pipeline.

Estimating the contribution of weather phenomena in an adverse image can help ease the model's ability to perform well in different weather conditions. In order to accomplish this, we leverage language guidance. We utilize the pretrained CLIP model, which learns a latent space that is shared by both image and text encodings~\cite{radford2021learning}. To formalize the problem, we are given an adverse image $\mathbf{I}$, which represents a clear image $\mathbf{J}$ which has been affected by some combination of functions $\mathcal{D}_{\text{rain}},\mathcal{D}_{\text{snow}},\mathcal{D}_{\text{fog}}$. For more information on the image formation model presented here, please refer to the supplementary materials. We also use a set of texts $\mathcal{T} = \{t_n\}_{n=1}^N$ describing $N$ different weather conditions, such as ``rain'', ``fog'', or ``snow''. We aim to find a vector ${\vec{v}} \in [0,1]^N$ where each element represents the contribution of each weather effect to an adverse image.

\begin{figure}[tb]
    \centering
    \includegraphics [width=\linewidth]{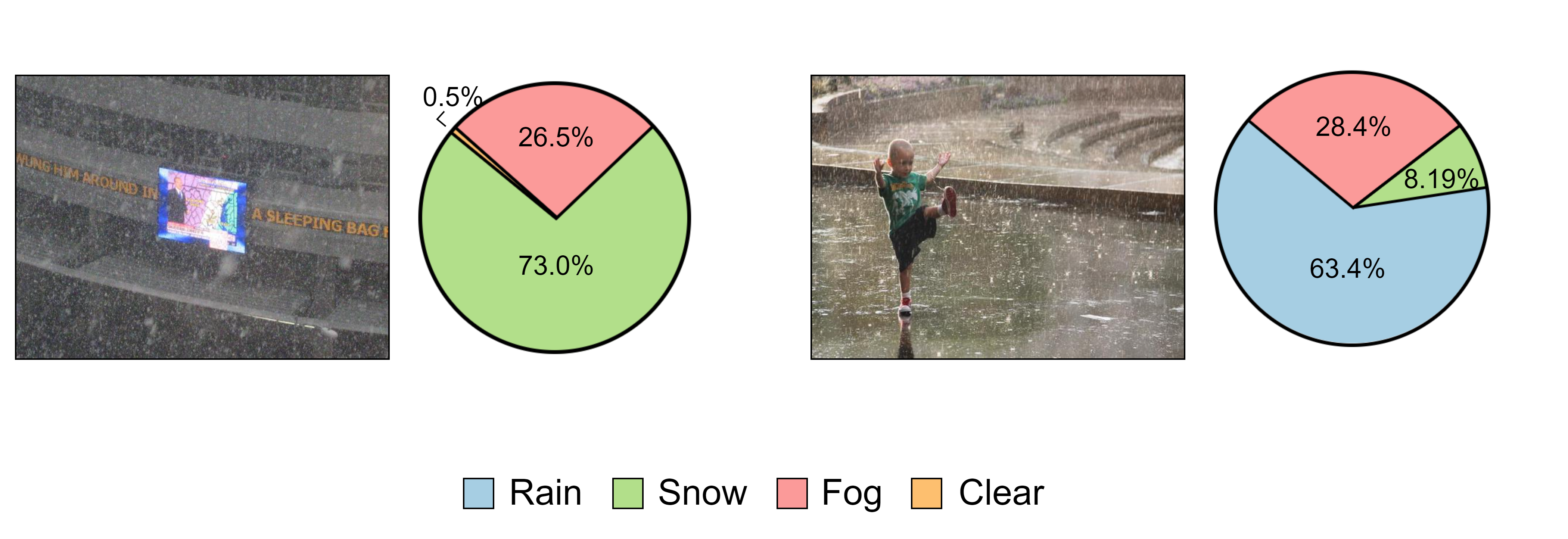}
    \caption{\textbf{Our CLIP injection layer is able to accurately predict the composition of weather effects in images.} The percentage of weather effect contributions was taken by passing in these images into our CLIP injection layer and extracting the weights $\vec{v}$.}
    \label{fig:clip_ref}
\end{figure}

We begin by passing $\mathbf{I}$ as well as each text $t_n$ through a frozen clip encoder model to obtain CLIP embeddings in the shared latent space:
\begin{align}
    \mathcal{C}_\mathbf{I} &= \text{CLIP}(\mathbf{I}), \\
    \mathcal{C}_\mathbf{T} &= \{\text{CLIP}(t_n)\}_{n=1}^N,
\end{align}
where the CLIP$()$ function represents passing an image or text through the CLIP encoder, $\mathcal{C}_\mathbf{I}$ denotes the length $512$ feature vector representing the adverse weather image $\mathbf{I}$, and $\mathcal{C}_\mathbf{T} \in \mathbb{R}^{N \times 512}$ denotes the matrix of CLIP feature vectors representing the set of weather texts $\mathcal{T}$. We pass $\mathcal{C}_\mathbf{I}$ through an MLP $f_\theta$ with parameters $\theta$ to obtain the $N$ length vector $\vec{v}$:
\begin{equation}
    \vec{v} = f_\theta(\mathcal{C}_\mathbf{I}).
\end{equation}
To learn the parameters $\theta$ such that an accurate weight vector $\vec{v}$ can be estimated, we perform a linear combination of text embeddings in the set $\mathcal{C}_\mathbf{T}$, weighted by $\vec{v}$, to obtain a final CLIP embedding $\vec{\mathcal{C}} \in \mathbb{R}^{512}$ representing the unique weather condition present in the adverse scene $\mathbf{I}$. This is concatenated with the CLIP embedding of the image $\mathcal{C}_\mathbf{I}$ to arrive at $\tilde{\vec{\mathcal{C}}} \in \mathbb{R}^{1024}$:
\begin{align}
    \vec{\mathcal{C}} &= \mathcal{C}_\mathbf{T} \cdot \vec{v}, \\
    \tilde{\vec{\mathcal{C}}} &= [\vec{\mathcal{C}} ; \mathcal{C}_\mathbf{I}],
\end{align}
where $[;]$ is the concatenation operation, $\tilde{\vec{\mathcal{C}}}$ is the final vector passed out of the layer into the rest of the model. To see examples of the learned weight vector $\vec{v}$ for unique scenes with different weather conditions, see~\cref{fig:clip_ref}.

\section{Experiments}
\label{sec:experiments}

\subsection{Implementation Details}
\label{sec:implementation}

For all model architectures that we trained and modified, we chose the best performing  configuration on Cityscapes that did not train with extra data. As such, we only used the default parameters for each model backbone, decode head, and auxiliary head. All three model architectures' default used UPerHead as their decode head and FCNHead as their auxiliary head in either their official repository or their official MMSegmentation~\cite{mmseg2020} implementation. For all models trained, we initialize each backbone with its respective generally pretrained checkpoint, which are trained on ImageNet. We use 20 text embeddings for the CLIP injection method. We evaluate our performance using the IoU and mIoU metrics. Hyperparameters, computational resources and definition of evaluation metrics can be found in the Supp. Mat.

\subsubsection{InternImage.}
We modified InternImage's~\cite{wang2023internimage} XL backbone by adding a cross-attention layer for CLIP Injections at the end of each stage. InternImage has a total of 39 layers in 4 stages, with stages ending at layers 5, 10, 34, and 39. Downsampling occurs after layers 5, 10, and 34. Before each downsampling layer of InternImage's encoder, a latent vector is outputted after a series of deformable convolution blocks, which is projected as queries and the CLIP injection $\tilde{\vec{\mathcal{C}}}$ is projected as the keys and values to the cross-attention layer. We add a total of 3 cross-attention layers to InternImage, which are added at the end of each stage except the last.

\subsubsection{ConvNeXt.}
ConvNeXt~\cite{liu2022convnet} has a total of 36 layers, with stages ending at layers 3, 6, 33, and 36. Downsampling occurs before layers 1, 3, 6, and 33. We add a cross-attention layer to ConvNeXt's XL backbone at the end of each stage except the last. In total, 3 cross-attention layers are added. We use their default batch size of 2 and learning rate of 8e-5.

\subsubsection{SWIN Transformer.}
We use Swin Transformer's~\cite{liu2021swin} config for their 12 window model from the official MMSegmentation~\cite{mmseg2020} implementation. We use their 12 window 22k pretrained model and maintain their default 24 layers and batch size of 2. We use a tuned learning rate of 6e-5 and, for our CLIP models, add the CLIP injection layers before each downsampling layer, which happens at the end of each stage excluding the last one. However, we also exclude the first stage from the CLIP injection, as Swin Transformer's architecture has a very large encoding at the end of the first stage. Therefore, adding a CLIP injection layer at that stage can greatly increase the memory usage above the capacity of most GPUs, which is why we exlcude it from that layer. Thus, we have 2 CLIP injection layers for our augmented Swin Transformer model. 

\subsection{Results}
\label{sec:results}
\subsubsection{WeatherProof.}
For the tests on our \dname, we follow the recommendations of the WeatherStream~\cite{zhang2023weatherstream} paper and use crop sizes of 224x224. We see in~\cref{table:Degraded-table} that our CLIP-guidance method for all model architectures results in improved performance by up to 10.2\% compared to the baseline models. This performance gap is likely due to the model's increased ability to handle more complex weather degradation patterns, such as those including multiple weather effects. Certain classes like stone end up performing much better, since smaller and less vibrant objects are more easily obfuscated by these complex weather patterns. By leveraging the rich latent space of CLIP to learn the composition of complex weather effects, the model is able to more easily segment and classify these objects. We additionally validate in~\cref{table:Clean-table} that our CLIP-guidance method does not harm (and in fact benefits in some cases) the model's performance on clear-weather data.


\begin{table}[tb]
 \caption{\textbf{On \dname, our proposed training method outperforms standard fine-tuning baselines for InternImage~\cite{wang2023internimage}, ConvNeXt~\cite{liu2022convnet}, and SWIN~\cite{liu2021swin, liu2022swin} when evaluating on adverse weather images.}}
  \label{table:Degraded-table}
  \scriptsize
  \centering
  \resizebox{\textwidth}{!}{
      \begin{tabular}{ccccccccccc}
        \toprule
        Model & \rotatebox{90}{Tree} & \rotatebox{90}{Struc.} & \rotatebox{90}{Road} & \rotatebox{90}{T-Snow} & \rotatebox{90}{T-Veg.} & \rotatebox{90}{T-Other} & \rotatebox{90}{Stone} & \rotatebox{90}{Building} & \rotatebox{90}{Sky} & mIoU $\uparrow$ \\
        \midrule
        InternImage~\cite{wang2023internimage} & \textbf{74.07} & 35.93 & \textbf{7.58} & 64.97 & \textbf{61.22} & 23.36 & 39.73 & 65.99 & 45.35 & 46.47\\
        InternImage~\cite{wang2023internimage} + Ours & 71.24 & \textbf{44.54} & 5.07 & \textbf{66.46} & 59.45 & \textbf{25.36} & \textbf{61.57} & \textbf{73.94} & \textbf{71.24} & \textbf{51.19}\\
        \midrule
        ConvNeXt~\cite{liu2022convnet} & 66.00 & \textbf{40.58} & 3.55 & \textbf{50.24} & 57.34 & \textbf{15.00} & 7.52 & 50.92 & \textbf{50.31} & 37.94\\
        ConvNeXt~\cite{liu2022convnet} + Ours & \textbf{68.98} & 32.58 & \textbf{5.70} & 44.51 & \textbf{60.52} & 11.75 & \textbf{30.48} & \textbf{62.50} & 42.27 & \textbf{39.92}\\
        \midrule
        SWIN~\cite{liu2021swin, liu2022swin} & \textbf{72.14} & 30.19 & 5.04 & \textbf{58.32} & \textbf{62.60} & 13.97 & 21.26 & \textbf{62.05} & \textbf{46.02} & 41.29\\
        SWIN~\cite{liu2021swin, liu2022swin} + Ours & 71.79 & \textbf{33.16} & \textbf{17.80} & 53.71 & 59.93 & \textbf{14.20} & \textbf{46.19} & 60.52 & 37.99 & \textbf{43.92}\\
        \bottomrule
      \end{tabular}
    }
\end{table}

\begin{table}[tb]
\caption{\textbf{On \dname, InternImage~\cite{wang2023internimage}, ConvNeXt~\cite{liu2022convnet}, and SWIN~\cite{liu2021swin, liu2022swin} still perform as well or better on clear images when using CLIP-based language guidance.}}
    \scriptsize
  \centering
  \resizebox{\textwidth}{!}{
      \begin{tabular}{ccccccccccc}
        \toprule
        Model & \rotatebox{90}{Tree} & \rotatebox{90}{Struc.} & \rotatebox{90}{Road} & \rotatebox{90}{T-Snow} & \rotatebox{90}{T-Veg.} & \rotatebox{90}{T-Other} & \rotatebox{90}{Stone} & \rotatebox{90}{Building} & \rotatebox{90}{Sky} & mIoU $\uparrow$ \\
        \midrule
        InternImage~\cite{wang2023internimage} & \textbf{78.78} & 40.49 & \textbf{9.77} & 67.97 & 62.39 & 23.57 & 62.08 & 67.97 & \textbf{76.97} & 54.44\\
        InternImage~\cite{wang2023internimage} + Ours & 77.74 & \textbf{47.13} & 9.74 & \textbf{70.34} & \textbf{64.14} & \textbf{28.29} & \textbf{70.42} & \textbf{73.78} & 72.64 & \textbf{57.14}\\
        \midrule
        ConvNeXt~\cite{liu2022convnet} & 72.85 & \textbf{48.09} & \textbf{5.29} & 54.37 & 60.94 & \textbf{20.32} & 20.03 & 56.14 & \textbf{76.81} & 46.09\\
        ConvNeXt~\cite{liu2022convnet} + Ours & \textbf{73.69} & 38.63 & 3.03 & \textbf{57.54} & \textbf{61.11} & 16.75 & \textbf{51.09} & \textbf{62.88} & 60.80 & \textbf{47.28}\\
        \midrule
        SWIN~\cite{liu2021swin, liu2022swin} & 75.14 & 35.77 & 7.37 & \textbf{62.83} & \textbf{62.85} & \textbf{18.67} & 33.06 & \textbf{63.00} & \textbf{68.53} & 47.47\\
        SWIN~\cite{liu2021swin, liu2022swin} + Ours & \textbf{77.44} & \textbf{39.74} & \textbf{29.12} & 58.39 & 61.25 & 16.95 & \textbf{75.14} & 62.46 & 65.17 & \textbf{53.96}\\
        \bottomrule
      \end{tabular}
  }
  \label{table:Clean-table}
\end{table}

\subsubsection{ACDC.}
The ACDC dataset~\cite{sakaridis2021acdc} is a semantic segmentation in adverse driving conditions dataset. It is comprised of 19 main classes that covers vehicles, structures, and pedestrians. The ACDC dataset includes snow, fog, rain, and night subcategories. Since night is out of the scope of this paper, we do not include it. We train using InternImage's default 512x1024 crop size for Cityscapes. Due to GPU memory constraints, we use a batch size of 1 and perform gradient accumulation every 2 training steps to get an effective batch size of 2, which is what InternImage uses by default. 
In ~\cref{table:ACDC-ablation}, we see that using the InternImage architecture along with CLIP language guidance results in an improved mIoU of 82.6. Compared to standard training techniques, our method performs 8.44\% better. Utilizing our language guidance led to improvements throughout all 19 classes. Specifically, objects such as trucks, buses, or motorcycles exhibited much larger performance increases since they are more heavily affected by complex weather effects such as rain and fog. By leveraging CLIP guidance to narrow the solution space, these objects become easier to identify.
In~\cref{table:ACDC-dataset}, we also see that using the InternImage architecture along with CLIP language guidance leads to a higher performance in rain, fog, and snow individually compared to previous SOTA models. Based on the average mIoU in rain, fog, and snow, our method is 6.21\% higher than the previous SOTA HRNet. We note that although the ACDC dataset is split into 3 different individual weather effect categories, it is still possible for there to be multiple weather effects like rain or snow present in a foggy scene. While we outperform HRNet in all 3 categories, we see the biggest jump in fog for this specific reason. Singular weather effects also still exhibit small performance gains, indicating that leveraging language guidance still helps due to the injection of rich latent information from CLIP. 
\begin{table}[tb]
\caption{\textbf{Our model, using language guidance, drastically improves performance compared to standard fine-tuning on the ACDC dataset.}}
  \centering
  \footnotesize
  \adjustbox{max width=\textwidth}{
      \begin{tabular}{c|cccccccccc}
        \toprule
        Model & road & sidew. & build. & wall & fence & pole & light & sign & veget. & terrain \\
        \midrule
        InternImage~\cite{wang2023internimage} & 94.69 & 80.81 & 90.81 & 68.94 & 60.46 & 66.10 & 78.93 & 75.37 & 91.86 & 67.03 \\
        InternImage~\cite{wang2023internimage} + Ours & \textbf{97.6} & \textbf{88.3} & \textbf{92.8} & \textbf{73.6} & \textbf{63.8} & \textbf{72.3} & \textbf{84.0} & \textbf{79.7} & \textbf{93.1} & \textbf{68.3} \\
        \midrule
        & sky & person & rider & car & truck & bus & train & motorc. & bicycle & \textbf{mIoU} $\uparrow$ \\
        \midrule
        InternImage~\cite{wang2023internimage} & 98.18 & 76.19 & 54.07 & 90.51 & 65.43 & 73.82 & 85.37 & 63.74 & 64.85 & 76.17 \\
        InternImage~\cite{wang2023internimage} + Ours & \textbf{98.6} & \textbf{81.6} & \textbf{62.5} &	\textbf{93.4} & \textbf{87.7} & \textbf{93.4} & \textbf{93.4} & \textbf{76.6} & \textbf{68.1} & \textbf{82.6} \\

        \bottomrule
      \end{tabular}
    }
  \label{table:ACDC-ablation}
\end{table}

\begin{table}[tb]
\caption{\textbf{Our language guided model achieves SOTA results on the ACDC dataset.} The Average mIoU is calculated by averaging between the three categories.}
  \centering
  \footnotesize
  \adjustbox{max width=\textwidth}{
      \begin{tabular}{c|cccc}
        \toprule
        Model & Rain mIoU & Fog mIoU & Snow mIoU & Average mIoU \\
        \midrule
        RefineNet~\cite{lin2017refinenet} & 68.7 & 65.7 & 65.9 & 66.9 \\
        DeepLabv2~\cite{chen2017deeplab} & 57.6 & 52.2 & 56.8 & 55.5 \\
        DeepLabv3+~\cite{chen2018encoder} & 74.1 & 69.1 & 69.6 & 70.9 \\
        HRNet~\cite{wang2020deep} & 77.7 & 74.7 & 76.3 & 76.2\\
        InternImage~\cite{wang2023internimage} & 75.18 & 78.54 & 72.10 & 75.27\\
        InternImage~\cite{wang2023internimage} + Ours & \textbf{79.4} & \textbf{83.3} & \textbf{80.2} & \textbf{81.0}\\
        \bottomrule
      \end{tabular}
    }
  \label{table:ACDC-dataset}
\end{table}

\subsubsection{Man-Made Adverse Conditions.}
The A2I2-Haze dataset~\cite{narayanan2023multi} is comprised of 14 classes including different categories of vehicles, objects, and humans. Images include man-made smoke and haze rather than natural weather phenomenon. It was originally labeled for the object detection task. For the following experiments, we converted the polygonal object detection masks to semantic segmentation labels. We train using crop size of 256x256. In~\cref{table:A2I2-dataset}, we see an increase of 3.9\% mIoU when training with the CLIP-based language guidance than without. This increase in performance suggests that the CLIP language guidance is also able to improve performance on non-natural man-made weather effects as well, such as smoke. This result indicates that leveraging the rich latent space of CLIP can help the model perform beyond natural weather effects. It also cements that the inclusion of language guidance helps with singular weather effects.

    
    

\begin{table}[tb]
\caption{\textbf{InternImage~\cite{wang2023internimage} performs better on the A2I2-Haze dataset when leveraging language guidance.} The use of CLIP-based guidance also helps models generalize beyond standard natural weather phenomenon to man-made smoke effects.}
  \centering
  \scriptsize
  \adjustbox{max width=\textwidth}{
  \begin{tabular}{c|cccccccc}
    \toprule
    Model & Case & Sedan & Pickup & Van & Tool & Ext. & Truss & \textbf{mIoU} $\uparrow$ \\
    \midrule
    InternImage~\cite{wang2023internimage} & \textbf{99.23} & \textbf{84.83} & 55.88 & 79.46 & 47.98 & 79.62 & 77.62 & 78.84 \\
    
    InternImage~\cite{wang2023internimage} + Ours & \textbf{99.23} & 78.78 & \textbf{70.03} & \textbf{88.18} & \textbf{53.49} & \textbf{87.55} & \textbf{80.33} & \textbf{81.89}\\

    \midrule
    & Barrel & Helmet & UTV & Barrier & Rope & Survivor & Backpack & \textbf{mIoU} $\uparrow$ \\

    \midrule
    InternImage~\cite{wang2023internimage} & 53.62 & \textbf{86.84} & 95.5 & 96.18 & 77.37 & 90.08 & 79.57 & 78.84 \\

    InternImage~\cite{wang2023internimage} + Ours & \textbf{56.10} & 85.93 & \textbf{98.21} & \textbf{96.35} & \textbf{79.88} & \textbf{90.29} & \textbf{82.15} & \textbf{81.89} \\
    \bottomrule
  \end{tabular}
  }
  \label{table:A2I2-dataset}
\end{table}

\subsection{Ablation Studies}
\label{sec:Ablation}
\noindent
\subsubsection{CLIP Injection Method.} In~\cref{table:clip-type}, we develop and compare the performance of a number of different CLIP injection methods. We first test MultiCLIP, in which $\mathcal{C}_\mathbf{T}$ is concatenated in the second dimension with $\mathcal{C}_\mathbf{M} \in \mathbb{R}^{N \times 512}$, which is created by repeating $\mathcal{C}_\mathbf{I}$ N times. The resulting matrix is $\tilde{\vec{\mathcal{C}}} \in \mathbb{R}^{N \times 1024}$, which is fed into the model. This CLIP injection method lowers mIoU performance by 2.4\% compared to our proposed method. This decrease in performance can likely be attributed to not constraining the final composition of weather effects by learning a relation between $\mathbf{I}$ and $\mathcal{C}_\mathbf{T}$. The other CLIP injection method is to use only a set of $N=4$ texts in the set $\mathcal{T}$ to describe the weather conditions. We find that, by limiting the size of $\mathcal{T}$, we decrease the CLIP injection layer's ability to boost model learning, leading to a lower mIoU performance by 0.89\%.

\noindent
\subsubsection{CLIP vs. Transient Attributes.} In~\cref{table:TA-ablation}, we provide a comparison between language guidance based on CLIP and language guidance based on the Transient Attributes paper~\cite{Laffont14}. Transient Attributes provides a dataset of images labeled with relevant weather conditions and then trains a regressor to predict these attributes from the new images. We use this regressor to replace CLIP by representing the scene attributes using the regressor's 40 length vector of scene attributes. We then project their vector with a linear layer to a 256-length vector which is passed into the cross-attention layer in the model. Overall, the CLIP-based approach performs better, achieving a 4.4\% improvement, suggesting that while simpler classifiers do garner some performance improvement, they do not get the same level of improvement as compared to the larger CLIP model. We hypothesize that for the CLIP-based guidance, the injection of a concatenation of the image encoding with the weighted combination of weather text encodings helps the model learn not only the weather composition, but also the complex interactions between the weather effects and the underlying scene. This level of general scene understanding is only possible with larger models like CLIP, that have been pre-trained on over 400 million image-text pairs.

\begin{table}[tb]
 \caption{\textbf{Ablation studies show that clip injection with 20 text prompts performs best as compared to other forms of CLIP Injection.}}
  \centering
  \resizebox{\textwidth}{!}{
      \begin{tabular}{c|cccccccccc}
        \toprule
        Model & \rotatebox{90}{Tree} & \rotatebox{90}{Struc.} & \rotatebox{90}{Road} & \rotatebox{90}{T-Snow} & \rotatebox{90}{T-Veg.} & \rotatebox{90}{T-Other} & \rotatebox{90}{Stone} & \rotatebox{90}{Building} & \rotatebox{90}{Sky} & mIoU $\uparrow$\\
        \midrule 
        InternImage~\cite{wang2023internimage} + MultiCLIP & 72.11 & 43.37 & 4.75 & \textbf{67.65} & \textbf{61.39} & 23.09 & 56.95 & \textbf{74.13} & 46.50 & 49.99\\		
        InternImage~\cite{wang2023internimage} + 4CLIP & \textbf{72.37} & 43.32 & \textbf{5.20} & 67.22 & 60.04 & \textbf{26.34} & 59.46 & 73.40 & 49.34 & 50.74 \\
        InternImage~\cite{wang2023internimage} + 20CLIP & 71.24 & \textbf{44.54} & 5.07 & 66.46 & 59.45 & 25.36 & \textbf{61.57} & 73.94 & \textbf{71.24} & \textbf{51.19}\\
        \bottomrule
      \end{tabular}
    }
  \label{table:clip-type}
\end{table}

\begin{table}[tb]
  \caption{\textbf{Ablation studies show that using our language guidance module built on CLIP performs the best compared to the baseline model with more parameters and using standard classifiers like Transient Attributes~\cite{Laffont14}.}}
  \centering
  \resizebox{\textwidth}{!}{
      \begin{tabular}{c|cccccccccc}
        \toprule
        Model & \rotatebox{90}{Tree} & \rotatebox{90}{Struc.} & \rotatebox{90}{Road} & \rotatebox{90}{T-Snow} & \rotatebox{90}{T-Veg.} & \rotatebox{90}{T-Other} & \rotatebox{90}{Stone} & \rotatebox{90}{Building} & \rotatebox{90}{Sky} & mIoU $\uparrow$\\
        \midrule 
        InternImage~\cite{wang2023internimage} + Transient Attributes & \textbf{71.69} & 41.47 & \textbf{5.69} & \textbf{66.63} & \textbf{61.41} & 23.47 & 53.95 & 71.97 & 44.88 & 49.02\\
        InternImage~\cite{wang2023internimage} + Extra Params & 70.68 & \textbf{46.81} & 9.11 & 56.34 & 59.82 & 15.59 & 54.09 & 65.96 & 47.53 & 47.32\\								
        InternImage~\cite{wang2023internimage} + Ours & 71.24 & 44.54 & 5.07 & 66.46 & 59.45 & \textbf{25.36} & \textbf{61.57} & \textbf{73.94} & \textbf{71.24} & \textbf{51.19}\\
        \bottomrule
      \end{tabular}
    }
  \label{table:TA-ablation}
\end{table}

\noindent
\subsubsection{Extra Parameters.}
In~\cref{table:TA-ablation}, we compare our model with language-guided cross-attention to a model with only self-attention. Our CLIP-guided model performs much better with a 8.18\% increase in performance compared to the self-attention model. This result demonstrates that our increase in performance compared to the baseline InternImage is not caused by an increase in the amount of parameters. Instead, our performance gain is due to our method of extracting weather features and scene understanding using CLIP.

\vspace{-10pt}
\section{Conclusion}
\vspace{-5pt}
In this paper, we investigate the performance gap of semantic segmentation models in adverse weather conditions, and propose a novel language-guided technique to reduce this gap. We note that while our model produces significant performance improvements, there are still limitations to the proposed method. Particularly strong weather effects that cause large occluded areas make it difficult to segment and classify the underlying scene. 

On a different note, while our technique is focused on a specific downstream task, we believe the general strategy of utilizing the latent knowledge of weather and other degradations learned by visual-language models (VLMs) can help build systems that are robust to both natural and man-made occlusions. We have proposed a new framework to introduce language as guidance. The approach is general and may not be limited to semantic segmentation. This is beyond the scope of this work, but we are hopeful of its extensions to other vision tasks. As VLMs \cite{radford2021learning,li2022blip,li2023blip} improve, methods following our framework may leverage future iterations of VLMs to provide guidance. 


We additionally hope that by introducing our \dname, we inspire further research into this specific field of semantic segmentation in adverse weather, with the eventual hope that future segmentation models can achieve ADE-20K or Cityscapes levels of accuracy despite the presence of adverse visual degradations such as weather.

\clearpage  

%
%
\bibliographystyle{splncs04}
\bibliography{main}

\renewcommand\thesection{\Alph{section}} 
\renewcommand\thesubsection{\thesection.\alph{subsection}} 
\renewcommand\thefigure{\Alph{figure}} 
\renewcommand\thetable{\Alph{table}} 

\newcommand{\tabnohref}[1]{Tab.~{\color{red}#1}} 
\newcommand{\fignohref}[1]{Fig.~{\color{red}#1}} 
\newcommand{\secnohref}[1]{Sec.~{\color{red}#1}} 
\newcommand{\eqnohref}[1]{Eq.~{\color{red}#1}} 
\newcommand{\linenohref}[1]{L#1} 
\newcommand{\cnohref}[1]{[{\color{green}#1}]} 

\setcounter{figure}{0}
\setcounter{section}{0}
\setcounter{table}{0}

\clearpage
\section*{Supplementary Contents}

This supplement is organized as follows:

\begin{itemize}
    \item \Cref{sec:analysis} shows some additional analysis results on multiple weather combinations.
    \item \Cref{sec:physics} gives some background on the image formation model for weather.
    \item \Cref{sec:implementation} gives some additional descriptions for the evaluation metrics and hyperparameters used for the experiments.
    \item \Cref{sec:failure} shows some failure modes and limitations for our training method.
    \item \Cref{sec:awss} shows some comparison metrics against domain adaptation methods on the ACDC dataset.
    \item \Cref{sec:qual} shows some qualitative results from the \dname\ test set.
    \item \Cref{sec:hrnet} shows some additional tests on the ACDC dataset.
\end{itemize}

\section{Analysis of Multiple Weather Effects}
\label{sec:analysis}

We show in \fignohref{2} and \secnohref{3.2} in the main paper that complex weather patterns with multiple weather effects such as rain+fog or snow+fog is more difficult for semantic segmentation models to predict from. In~\Cref{table:multi-table}, we show this with quantitative analysis on the~\dname. We categorize the images in our dataset into either single or multi-effect weather patterns and test the InternImage model on both categories. The implementation details are the same as previous experiments with the model in \secnohref{5}. We first see that there is a substantial performance drop of around 14.9\% for multi-effect weather patterns as compared to single-effect patterns. Furthermore, by leveraging our CLIP-based language guidance, this gap is drastically reduced to around 3.5\%. 

\begin{table}[th]
 \caption{On \dname, we see that baseline InternImage performs worse on images with multiple weather effects. This gap shrinks once the CLIP-based language guidance is introduced.}
  \label{table:multi-table}
  \scriptsize
  \centering
  \resizebox{\textwidth}{!}{
      \begin{tabular}{cccccccccccc}
        \toprule
        & Method & \rotatebox{90}{Tree} & \rotatebox{90}{Struc.} & \rotatebox{90}{Road} & \rotatebox{90}{T-Snow} & \rotatebox{90}{T-Veg.} & \rotatebox{90}{T-Other} & \rotatebox{90}{Stone} & \rotatebox{90}{Building} & \rotatebox{90}{Sky} & mIoU $\uparrow$ \\
        \midrule
        \multirow{4}{*}{\rotatebox[origin=c]{90}{Single}} & InternImage~\cite{wang2023internimage} & \textbf{25.53} & 39.74 & \textbf{5.49} & \textbf{67.67} & 64.16 & 40.01 & 54.06 & 78.28 & \textbf{79.19} & 50.46\\ \\
        & InternImage~\cite{wang2023internimage} + Ours & 24.74 & \textbf{40.47} & 5.36 & 65.36 & \textbf{64.79} & \textbf{41.95} & \textbf{59.35} & \textbf{81.24} & 74.36 & \textbf{50.85}\\ \\
        \midrule
        \multirow{4}{*}{\rotatebox[origin=c]{90}{Multi}} & InternImage~\cite{wang2023internimage} & 81.51 & 38.73 & \textbf{3.51} & 65.8 & \textbf{58.17} & \textbf{10.56} & 36.87 & 61.28 & 30.05 & 42.94\\ \\
        & InternImage~\cite{wang2023internimage} + Ours & \textbf{81.52} & \textbf{48.34} & 2.51 & \textbf{67.14} & 54.87 & 9.39 & \textbf{68.27} & \textbf{67.18} & \textbf{42.57} & \textbf{49.09}\\ \\
        \bottomrule
      \end{tabular}
    }
\end{table}

\section{Image Formation Model}
\label{sec:physics}

In order to study and alleviate the performance gap faced by semantic segmentation models in weather conditions such as rain, fog, or snow, it is important for us to mathematically formalize how an image can be affected by different weather phenomenon. To do this, we initially borrow the light transport model proposed by research in the field of weather removal~\cite{ba2022not,zhang2023weatherstream,deng2018directional,fu2017removing,li2018non,li2019heavy,li2020all,li2016rain,valanarasu2022transweather,wang2020model,wang2019spatial,yasarla2019uncertainty,zhang2018density,zhu2017joint,chen2020jstasr,chen2021all,liu2018desnownet,he2010single,tan2008visibility}. Weather in an image can largely be attributed to one of two effects, particle effects (raindrops or snowflakes) or scattering effects (rain accumulation, snow veiling, haze, fog, etc.). Weather particles can be modeled as a convex combination of the underlying clear scene and a map of the particles. This can be done for rain as follows:
\begin{equation} \label{eq:rain_model}
    \mathcal{D}_{\text{rain}}(\mathbf{J}(x)) = \mathbf{J}(x)(1-\mathbf{M}_r(x)) + \mathbf{R}(x)\mathbf{M}_r(x),
\end{equation}
where $x$ represents the spatial location within an image, $\mathcal{D}_{\text{rain}}$ represents a function that maps a clear image to one with rain particle effects, $\mathbf{J}(x)$ represents the clear image with no weather effects, $\mathbf{M}_r(x)$ represents a mask of the locations of rain particles, and $\mathbf{R}(x)$ represents a map of the rain streaks~\cite{ba2022not,zhang2023weatherstream,deng2018directional,fu2017removing,li2018non,li2019heavy,li2020all,li2016rain,valanarasu2022transweather,wang2020model,wang2019spatial,yasarla2019uncertainty,zhang2018density,zhu2017joint}. For snow, it is:
\begin{equation} \label{eq:snow_model}
    \mathcal{D}_{\text{snow}}(\mathbf{J(x)}) = \mathbf{J}(x)(1-\mathbf{M}_s(x)) + \mathbf{S}(x)\mathbf{M}_s(x),
\end{equation}
where $\mathcal{D}_{\text{snow}}$ and $\mathbf{M}_s$ represent their corresponding snow equivalents, and $\mathbf{S}(x)$ represents a chromatic aberration map of the snow particles~\cite{chen2020jstasr,chen2021all,liu2018desnownet}.

Scattering effects are modeled through the use of the scene radiance equation, which, evaluated at each pixel location, is
\begin{equation} \label{eq:radiance}
\begin{split}
    \mathcal{D}_{\text{fog}}(\mathbf{J(x))} & = \mathbf{J}(x)e^{-\int_0^{d(x)}\beta dl} + \int_0^{d(x)}L_\infty \beta e^{-\beta l}dl, \\
    & = \mathbf{J}(x)e^{-\beta d(x)} + L_\infty(1-e^{-\beta d(x)}),
\end{split}
\end{equation}
where $\mathcal{D}_{\text{fog}}$ represents a function mapping a clear image to one with scattering effects, $d(x)$ represents the distance from the observer at a pixel location $x$, $\mathbf{J}(x)$ represents the radiance of the underlying scene (the clear image), $\beta$ is an atmospheric attenuation coefficient (assumed to be constant throughout the scene), and $L_\infty$ is the radiance of the airlight~\cite{tan2008visibility,he2010single}. 

Images degraded by adverse weather can be affected by any combination of~\cref{eq:rain_model,eq:snow_model,eq:radiance}. Having an estimate of how much an image is affected by each of these particular weather phenomenons can be utilized by the model to limit the search space of possibly advantageous feature representations. This estimation process is explained in~\secnohref{4}.

\section{Additional Implementation Details}
\label{sec:implementation}

We trained and tested our experiments using one RTX 3090. As discussed in \cref{sec:implementation}, we use mostly the same parameters as each model we test, only changing learning rate. For our CLIP models, we use 20 text embeddings covering rain, snow, fog, and clear weather descriptions.

The metric used is mean intersection over union (mIoU). Following the MMSegmentation~\cite{mmseg2020} class's implementation, we do the sum of all intersections of a particular class over the entire test set divided by the sum of all unions of a particular class over the entire test set. We then average together all of the classes in the dataset.

As shown in \cref{table:params}, the addition of the CLIP language guidance module leads to minimal increases in computational complexity and parameters of each model. For instance, InternImage has a 10.3\% increase in FLOPS and a 4.97\% increase in parameters. This is relatively efficient compared to the 10.2\% increase in performance on our \dname.

\begin{table}
  \caption{\textbf{Comparison of the FLOPS and parameters of the original models vs. our models.}}
  \centering
  \begin{tabular}{c|cc}
    \toprule
    Model & FLOPS (G) & Parameters (M) \\
    \midrule
    InternImage~\cite{wang2023internimage} & 95.66 & 367.62\\
    InternImage~\cite{wang2023internimage} w/ CLIP (Ours) & 105.56 & 385.90 \\
    \midrule
    ConvNeXt~\cite{liu2022convnet} & 101.95 & 391.03 \\
    ConvNeXt~\cite{liu2022convnet} w/ CLIP & 117.25 & 422.27 \\
    \midrule
    Swin~\cite{liu2021swin} & 64.07 & 121.42 \\
    Swin~\cite{liu2021swin} w/ CLIP & 69.08 & 129.57 \\
    \bottomrule
  \end{tabular}
  \label{table:params}
\end{table}

\section{Failure Modes and Limitations}
\label{sec:failure}
\begin{figure*}[t]
    \centering
    \includegraphics [width=\linewidth]{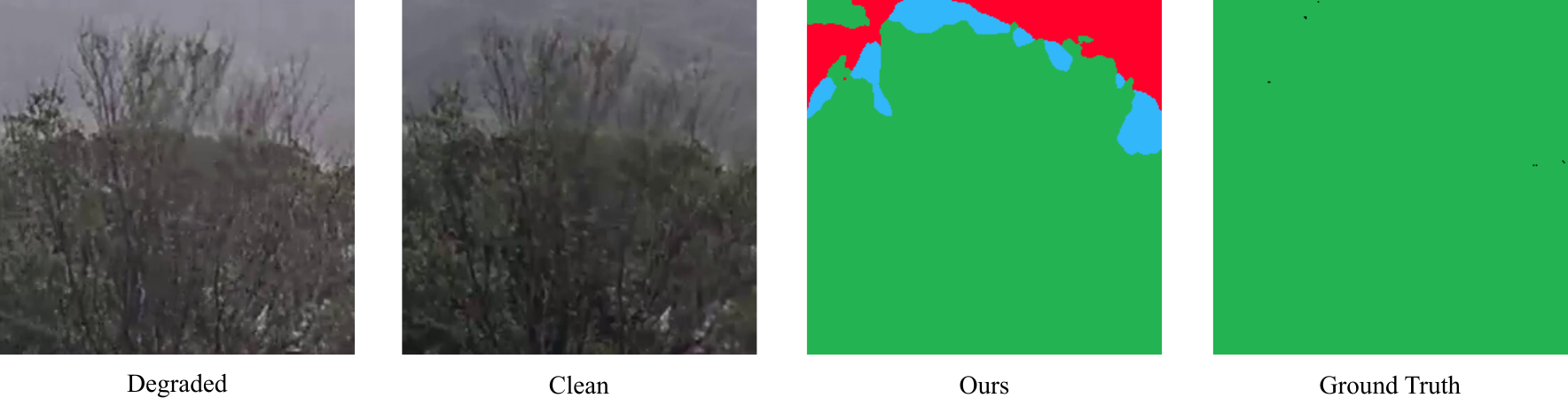}
    \caption{\textbf{Our CLIP-guided method does not perform well when there are occlusions that completely cover significant parts of a scene.} As shown in the figure, the background becomes masked from fog in the degraded image. As such, when training on an image like this, it is harder to recover details of the underlying scene.}
    \label{fig:failures}
\end{figure*}

As seen in~\Cref{fig:failures}, our CLIP guidance fails when the weather effects in an image are so significant that they fully obstruct parts of a scene. While this may be obvious for the shown scene, it is much harder to notice for weather effects that only show up in certain frames, such as large snow flakes that are close to the lens.

\section{Domain Adaptation Method Comparisons}
\label{sec:awss}\

In \cref{table:ACDC-Domain-Adaptation}, we train a model with CLIP guidance using the SOTA domain adaptation method for Cityscapes to ACDC. Comparing against Rein~\cite{wei2023stronger}, we gain a performance increase of 1.3\% on average mIoU. Like in the previous table, night scenes perform worse as they are out of the purview of this paper. However, our method still outperforms the current SOTA for ACDC.

\begin{table}[tb]
\caption{The results of using domain adaptation methods on the rain, fog, and snow scenes of the ACDC val dataset. The Average mIoU is calculated by averaging between the three categories.}
  \centering
  \footnotesize
  \adjustbox{max width=\textwidth}{
      \begin{tabular}{c|cccc}
        \toprule
        Model & Rain mIoU & Fog mIoU & Snow mIoU & Average mIoU \\
        \midrule
        Rein~\cite{wei2023stronger} & 78.2 & 76.4 & \textbf{79.5} & 78.0 \\
        Rein~\cite{wei2023stronger} + Ours & \textbf{78.4} & \textbf{82.3} & 76.2 & \textbf{79.0}\\
        \bottomrule
      \end{tabular}
    }
  \label{table:ACDC-Domain-Adaptation}
\end{table}

\section{Qualitative Results}
\label{sec:qual}

\begin{figure*}[t]
    \centering
    \includegraphics [width=\linewidth]{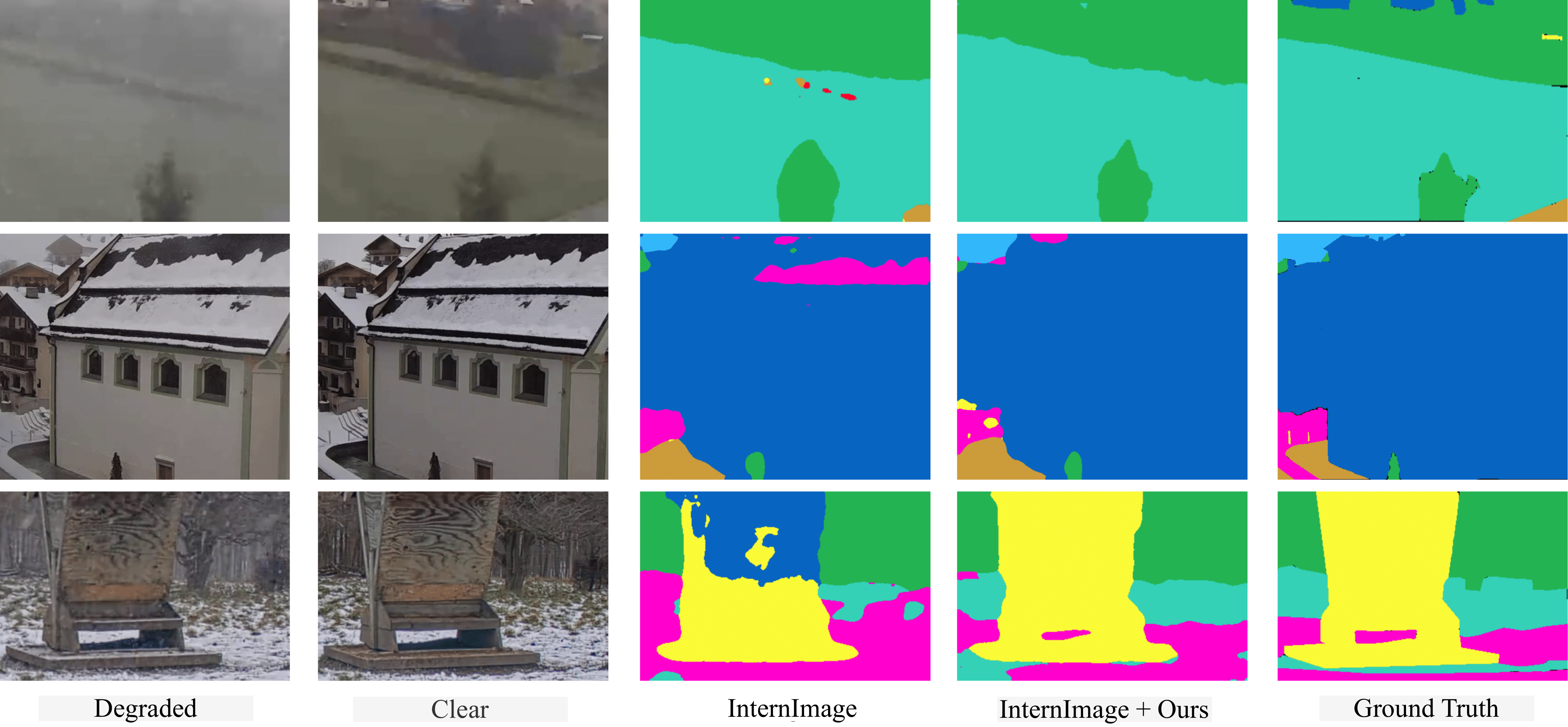}
    \caption{\textbf{Qualitative results from InternImage trained with with and without our proposed CLIP guidance.} The degraded image, clean image, and ground truth semantic segmentation maps are also included for reference. MMsegmentation's decoders produce a segmentation map that is 0.25 the size of the original image and use bilinear interpolation to get the full map. Thus, both models often lack the ability to produce high frequency details and have much smoother labels.}
    \label{fig:qualititative}
\end{figure*}

In~\Cref{fig:qualititative}, we show the qualitative results of our models compared to the baseline model. This comparison was done on the InternImage model. MMSegmentation~\cite{mmseg2020} upscales by 4, meaning many high frequency details are not present in the outputs.

\section{Additional Experiments on ACDC}
\label{sec:hrnet}

In \cref{table:ACDC-dataset-extra}, we compare against SOTA models for semenatic segmentation on ACDC. Unlike the ACDC results in \secnohref{5.2} of the main paper, we also train and test on night scenes here. We include the results below to show the effects of including nighttime images in the training and testing set and comparing against the full ACDC test set. Note that our method is limited to the weather effects of ``rain'', ``fog'', and ``snow'', and so nighttime images are out of the scope of this paper. As expected, the results in the night category are worse than HRNet. However, on average, our method still outperforms the current SOTA for ACDC by 1.8\% even while including the disadvantageous nighttime scenes. We also note that this work into language can be extended in the future for multiple different degradations, and is not limited to weather. We leave the application of language guidance to low-light nighttime images to future work.

\begin{table}[tb]
\caption{The results below are on the full ACDC val dataset, including night-time images. The Average mIoU is calculated by averaging between the four categories.}
  \centering
  \footnotesize
  \adjustbox{max width=\textwidth}{
      \begin{tabular}{c|ccccc}
        \toprule
        Model & Night mIoU & Rain mIoU & Fog mIoU & Snow mIoU & Average mIoU \\
        \midrule
        RefineNet~\cite{lin2017refinenet} & 55.5 & 68.7 & 65.7 & 65.9 & 64.0 \\
        DeepLabv2~\cite{chen2017deeplab} & 45.3 & 57.6 & 52.2 & 56.8 & 53.0 \\
        DeepLabv3+~\cite{chen2018encoder} & 60.9 & 74.1 & 69.1 & 69.6 & 68.4 \\
        HRNet~\cite{wang2020deep} & \textbf{65.3} & \textbf{77.7} & 74.7 & 76.3 & 73.5\\
        InternImage~\cite{wang2023internimage} + Ours & 59.3 & 77.1 & \textbf{83.2} & \textbf{79.5} & \textbf{74.8}\\
        \bottomrule
      \end{tabular}
    }
  \label{table:ACDC-dataset-extra}
\end{table}
\end{document}